# Towards Deep Industrial Transfer Learning for Anomaly Detection on Time Series Data

Benjamin Maschler
Institute of Industrial Automation and
Software Engineering
University of Stuttgart
Stuttgart, Germany
benjamin.maschler@ias.uni-stuttgart.de

Tim Knodel
Institute of Industrial Automation and
Software Engineering
University of Stuttgart
Stuttgart, Germany

Michael Weyrich
Institute of Industrial Automation and
Software Engineering
University of Stuttgart
Stuttgart, Germany
michael.weyrich@ias.uni-stuttgart.de

*Abstract*— Deep learning promises performant anomaly detection on time-variant datasets, but greatly suffers from low availability of suitable training datasets and frequently changing tasks. Deep transfer learning offers mitigation by letting algorithms built upon previous knowledge from different tasks or locations. In this article, a modular deep learning algorithm for anomaly detection on time series datasets is presented that allows for an easy integration of such transfer learning capabilities. It is thoroughly tested on a dataset from a discrete manufacturing process in order to prove its fundamental adequacy towards deep industrial transfer learning – the transfer of knowledge in industrial applications' special environment.

## I. Introduction

In all automated systems, anomalies pose a challenge to any control software and are therefore considered to be a potential problem. They are difficult to handle even in data-driven systems, because, by definition, they are what is not addressed by conventional rule or model based automation and might occur unexpectedly as well as differ considerably from any previous occurrence [1–3]. Simply ignoring them can be detrimental with consequences ranging from inefficiencies [4] to complete failure [5], possibly harming workers or users [6].

Handling anomalies starts with detecting them. Here, a shift from conventional, static methods towards deep learning based, dynamic approaches could be witnessed in recent years [1, 2]. Nevertheless, data scarcity and high process dynamics remain challenging [7–9].

Mitigation could be provided by knowledge transfer between detection algorithms, bridging over gaps between different smaller datasets or various states of a process. This could be realized using transfer learning approaches, which allow the use of previously acquired knowledge for improving the learning of new tasks [10].

On the task of image recognition, this approach yielded exceptional results in recent years [7]. However, in industry, time series data is prevalent, calling for different or at least adapted approaches in order to realize similar gains.

*Objective*: In this article, we present a deep-learning-based, modular anomaly detection algorithm for time series data. It features characteristics that allow for an easy integration of transfer learning capabilities. An evaluation is carried out on a discrete manufacturing process.

*Structure*: In chapter II, related work on anomaly detection using deep learning and the challenges therein is presented. Transfer learning is introduced, offering mitigation. Chapter III describes a modular deep-learning-based anomaly detection algorithm. It is evaluated in chapter V. Chapter VI then proposes a deep industrial transfer learning architecture building upon the modular, but otherwise conventional approach presented before. Finally, chapter VII presents a conclusion, underlining the concepts fundamental feasibility to transfer-learning-based anomaly detection on time series data.

## II. Related Work

In this chapter, a brief overview of anomaly detection using deep learning methods is presented. The challenges that hinder a widespread utilization of such methods are discussed, leading to the introduction of the concepts of transfer and continual learning as a basis for mitigation.

### A. Anomaly Detection using Deep Learning

In technical systems, anomalies are defined as irregularities which are not considered to be part of a system's normal, intended behavior. Because of their oftentimes unknown origin and inadvertent occurrence, anomalous dynamics lead to instabilities and thereby cause increased inefficiencies and system errors [1–3].

Despite some variance in terminology in literature, there are usually three different types of anomalies (see Fig. 1) differentiated:

*Point anomalies* are characterized by a single data instance being considered anomalous. They are usually the easiest to detect and have been subject to extensive research [1–3].

*Collective anomalies* (or group anomalies) are characterized by a series of data instances being considered anomalous as a group, although not each individual data instance needs to be anomalous. This necessarily requires relations between individual data instances, such as their sequence in time series data [1–3].

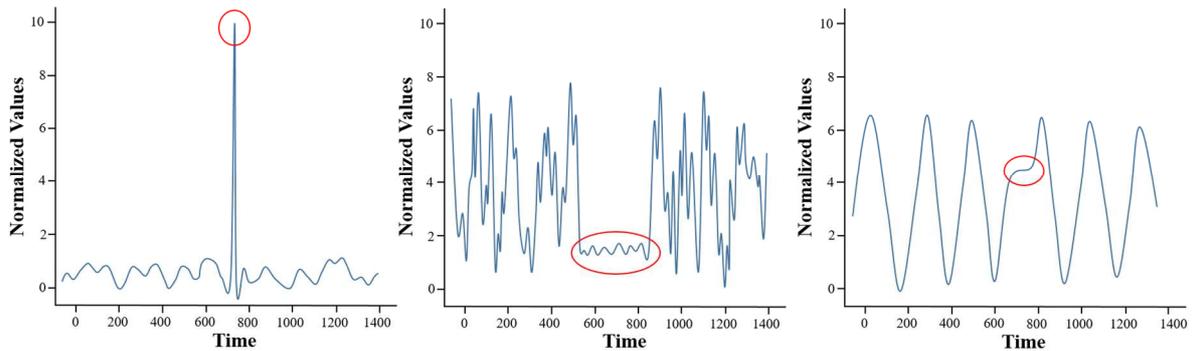

Fig. 1. Point anomaly (left), collective anomaly (middle) and contextual anomaly (right) according to [3]

*Contextual anomalies* (or conditional anomalies) are characterized by data being anomalous in specific, separately defined contexts or conditions only. This necessarily requires contextual attributes, normally regarding at least time or space, describing the data instances' context [1–3].

Traditionally, statistical, classification-based, clustering-based or information theoretic approaches are used for detecting anomalies. A shared characteristic of those approaches is their limitation on detecting anomalies based on static, time-invariant models only. Due to the fact that many anomalies, especially collective and contextual ones, are dynamic and time-variant, different approaches are needed for a more robust and widely applicable anomaly detection [2, 3].

Here, deep-learning-based approaches have started to present an alternative. Especially recurrent neural networks, e.g. long short-term memory (LSTM) and autoencoders (AE), have demonstrated their capability of detecting anomalies in dynamic and time-variant scenarios [2]:

In [6], a variational AE is used for anomaly detection based upon a comparison of log-likelihood real and reconstructed data. The algorithm outperforms state-of-the-art detectors on a dataset collected from a care robot tasked with feeding different people.

In [4], a stacked AE with interconnected gated recurrent units (GRU) for short-term and LSTM for long-term memory is used to detect anomalies as well as to predict the process outputs. The algorithm uniquely performs those different tasks simultaneously on a dataset collected from a multi-step hot forging process on hydraulic presses.

In [5], LSTM are combined with exponentially weighted moving average (EWMA) in order to significantly increase efficiency on contextual anomaly detection. This is demonstrated on a dataset collected from two simulated industrial robotic manipulators collaborating.

This demonstrates deep-learning-based approaches' fundamental capability for solving real-life industrial anomaly detection problems. However, limitations caused by scarce training data on dynamic and fragmented scenarios still remain a major challenge to their practical application.

### B. Deep Industrial Transfer Learning

The term 'deep industrial transfer learning' refers to methods utilizing previously acquired knowledge within deep learning techniques to solve tasks from the industrial domain [7, 10]. It can be used to facilitate learning across several smaller, less homogenous datasets [11–13], thereby mitigating two central problems of conventional deep learning in industry [10]:

- Machinery and processes are often unique or at least heavily customized. Together with high data protection levels and strong competition among different entities [14], the acquisition of large and diverse datasets ranges between difficult and outright impossible [8, 9].

- Economic volatility, shorter innovation and product life cycles are driving reconfiguration demand [15, 16]. Coupled with wear and changing environments, this renders most tasks highly dynamic. Datasets once acquired only prove a short-term representation of a problem space, necessitating continuous data collection and algorithm retraining [7].

*Continual learning*, i.e. the transfer of knowledge and skills from one or more source tasks to a target task in order to solve source and target tasks [17, 18], promises to also address and possibly solve this problem. Among the three methodological classes within continual learning, only one has been applied to industrial use cases: regularization methods. However, recent studies have raised concerns regarding the robustness and scalability of such approaches:

In [11], elastic weight consolidation (EWC) [19] is used to predict the state-of-health of turbofan engines. The dataset used was collected through simulation and although the different sub-problems were very similar, the algorithm struggled with generalization depending on how similar they were.

In [13], EWC, online EWC [20] and synaptic intelligence (SI) [21] are compared regarding their ability to predict the state-of-health of lithium-ion batteries. Despite the overall good results, again, the generalization capabilities showed clear limitations.

In [12], EWC, online EWC, SI and learning without forgetting (LwF) [22] are compared regarding their ability to detect anomalies in a discrete manufacturing process. Here, a saturation-like effect is limiting the continued addition of new knowledge for longer task sequences, raising concerns regarding the approaches' long-time functionality.

Because of these limitations, *transfer learning*, i.e. the transfer of knowledge and skills from one or more source tasks to a target task in order to improve solving the target task [10, 23, 24], might be better suited to mitigate the aforementioned problems. Indeed, e.g. on image recognition, it enabled the well-known performance gains witnessed in recent years by allowing pre-trained feature extraction algorithms to be re-trained on previously unseen objects [7]. Thereby, even one-shot learning on single-chip computers became possible [25]. For industrial applications, however, similar improvements are necessary regarding leaning on time series data.

### III. MODULAR DEEP-LEARNING-BASED ANOMALY DETECTION ALGORITHM

Our concept is loosely based on the dual-memory method, which uses a fast-learning and a slow-learning sub-system to better address the dilemma of having to retain important information potentially permanently while still being able to take up new knowledge [17, 26]. However, we aim to adapt it from a continual learning approach aimed at solving old tasks as well as new sub-tasks to a transfer learning approach focused only on the sub-task at hand.

The basis for such a dual-memory transfer learning algorithm is a modular, but otherwise conventional learning algorithm, here: a modular deep-learning-based anomaly detection algorithm (see Fig. 2). It does not have transfer learning capabilities, yet, but is the necessary foundation for such an extension. Furthermore, it allows a validation of the input and output modules which are to be used in the transfer learning algorithm as well.

#### A. Input Module - Feature Extractor

A slow-learning (in our case: static) module is supposed to ensure generalization across different (sub-)tasks. It serves as the *input module* to the overall algorithm and is a pre-trained feature extractor (see Fig. 2, letter A). It compresses the input data, thereby greatly reducing storage and computing resource requirements.

In order to find a suitable basis for the later to be static feature extractor, three different architectures were set up. All of them are encoding parts of AEs which are capable of robust feature extraction. In training, the Adam optimizer was used on batch sizes of 32 samples over 10 epochs with losses defined as the mean squared error (MSE).

The first approach is based upon a stacked LSTM network for unsupervised feature extraction. In literature, it demonstrated a good performance on industrial times series data [27]. Here, two layers of LSTM cells compress the input vector to the chosen code size.

The second approach is based upon a one-dimensional convolutional neural network (CNN). In literature, it, too, demonstrated a good performance on industrial time series data [28]. Here, three convolutional layers are used to compress the input vector to the chosen code size.

The third approach is based upon a fully connected (FC) neural network. It is the simplest form of an autoencoder and was therefore chosen as a baseline comparison. Two layers were used to compress the input vector to the chosen code size.

Table I lists the hyperparameters used for the experiments in chapter IV. They were chosen based upon published literature and previous experience. Due to time constraints, no extensive hyperparameter optimization was conducted.

#### B. Output Module – Anomaly Detector

The fast-learning module is supposed to ensure differentiation between different sub-tasks. It serves as an *output module* to the overall algorithm and uses previously extracted feature vectors as an input (see Fig. 2, letter B).

In order to find a suitable basis for the easily retrainable anomaly detector, another FC AE was chosen for simplicity. By further compressing the feature vector to half its size and comparing the resulting reconstruction to the original, it is capable of finding anomalies when trained with normal samples only.

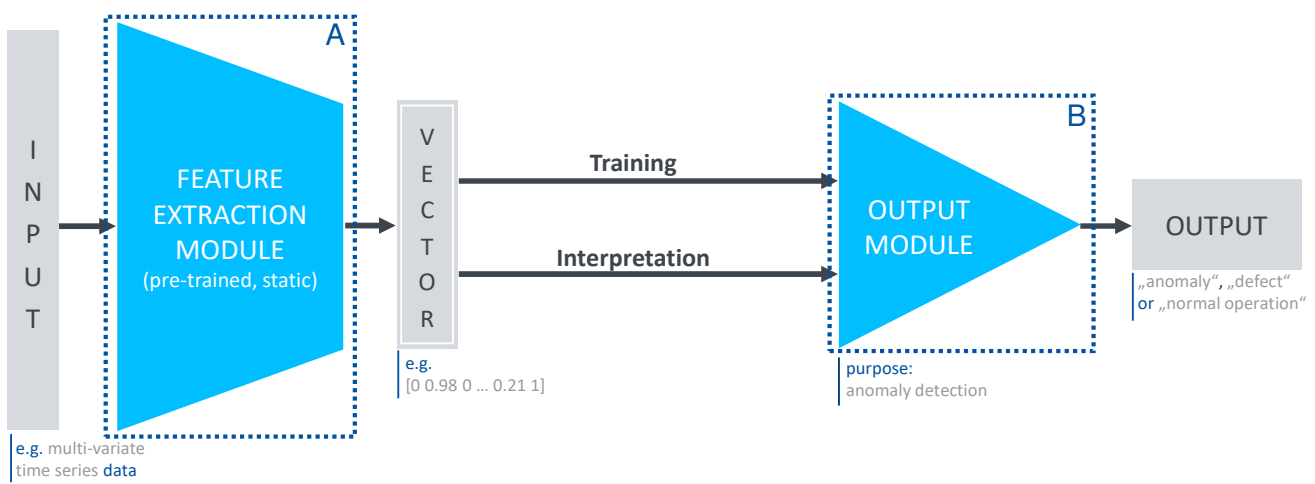

Fig. 2. Modular deep-learning-based anomaly detection algorithm

TABLE I. HYPERPARAMETERS OF THE FEATURE EXTRACTING INPUT MODULE

| Architecture | Parameter | Value |
|---|---|---|
| all | Input size | 3,000 |
| | Batch size | 32 |
| | Optimizer | Adam |
| | Max. epochs | 10 |
| | Loss function | MSE |
| LSTM | Hidden size (layer 1) | <code size> * 10 |
| | Hidden size (layer 2) | <code size> |
| CNN | Kernel (layer 1) | 10 |
| | Stride (layer 1) | 5 |
| | Activation function (layer 1) | ReLU |
| | Kernel (layer 2) | 5 |
| | Stride (layer 2) | 3 |
| | Input size (layer 2) | <code size> / 3 |
| | Activation function (layer 2) | ReLU |
| | Kernel (layer 3) | 3 |
| | Stride (layer 3) | 2 |
| | Input size (layer 3) | <code size> / 2 |
| | Output channel (layer 3) | <code size> |
| | Activation function (layer 3) | SELU |
| FC | Number of nodes (layer 1) | 500 |
| | Activation function (layer 1) | ReLU |
| | Number of nodes (layer 2) | <code size> |
| | Activation function (layer 2) | SELU |

TABLE II. HYPERPARAMETERS OF THE ANOMALY DETECTING OUTPUT MODULE

| Parameter | Value |
|---|---|
| Input size | <code size> |
| Batch size | 32 |
| Optimizer | Adam |
| Max. epochs | 10 |
| Loss function | MSE |
| Number of nodes (encoder) | <code size> / 2 |
| Activation function (encoder 1) | ReLU |
| Number of nodes (decoder) | <code size> |
| Activation function (decoder) | SELU |

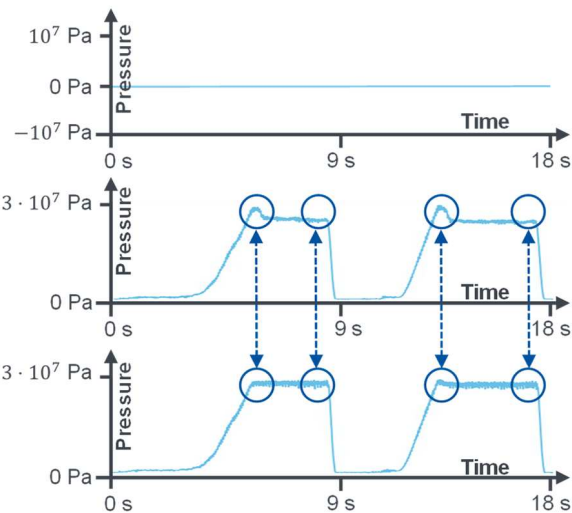

Fig. 3. Comparison of pressure data from defect (top), anomalous (middle) and normal (bottom) pumps (main deviations highlighted)

Table II lists the hyperparameters used for the experiments in chapter IV.C and IV.D. They were chosen based upon published literature and previous experience. Due to time constraints, no extensive hyperparameter optimization was conducted.

IV. EVALUATION

In this chapter, an evaluation of the modular deep-learning-based anomaly detection algorithm is carried out. First, the dataset used is characterized, before the algorithm's performance in different experiments is detailed.

All experiments were conducted on a computer featuring an AMD Ryzen Threadripper 2920X CPU and a NVIDIA GeForce RTX 2080 8 GB GPU running Ubuntu 20.04. The learning framework used was PyTorch 1.6 under Python 3.6.

A. Experimental Dataset

The experiments were conducted using the subset of a very large industrial metal forming dataset collected on a hydraulic press. It consists of data from eight pumps applying pressure on a shared oil reservoir. Du to this setup, anomalous behavior of one pump is compensated by other pumps. This hides such behavior from the operator, because initially no problems occur. However, the other pumps experience increased wear, which makes an early detection of the described anomalous behavior desirable.

Apart from data on normal and anomalous pumps, the dataset contains measurements of defect pumps. Because of their completely different sensor readings they are easy to identify automatically (see Fig. 3, top).

However, the differentiation between normal and anomalous pump behavior is much harder (see Fig. 3, middle and bottom). The challenge is further increased by frequent alterations of the production process, either by improvements such as new molds or changes of the manufactured product. Every alteration causes a change of the process' characteristics, which requires anomaly detection algorithms to be retrained. Here, deep industrial transfer learning could greatly reduce the computational effort and the amount of data needed.

B. Input Module – Feature Extractor

For an initial evaluation of the feature extracting input module, different versions of it are trained and tested. Apart from the different deep learning methods LSTM, CNN and FC, the impact of different code sizes on the reconstruction performance was analyzed: The algorithm compressed the initial 3,000 value input vector into a vector of length 50, 100 or 150.

For this experiment's training, unlabeled pressure data from the production of nine different products, each being produced thousands of times, is used. After excluding data from defect pumps, an average of all remaining pumps for each production event is calculated and used as input.

TABLE IV. DIFFERENT FEATURE EXTRACTING INPUT MODULES' TEST LOSSES

| Architecture | Code size | Test loss |
|---|---|---|
| LSTM | 50 | 4.71 x 10$^{-3}$ |
|  | 100 | 4.05 x 10$^{-3}$ |
|  | 150 | **3.91 x 10$^{-3}$** |
| CNN | 50 | 1.17 x 10$^{-4}$ |
|  | 100 | 1.09 x 10$^{-4}$ |
|  | 150 | **2.73 x 10$^{-5}$** |
| FC | 50 | **1.96 x 10$^{-3}$** |
|  | 100 | 2.35 x 10$^{-3}$ |
|  | 150 | 2.66 x 10$^{-3}$ |

bold: architectures' best accuracy

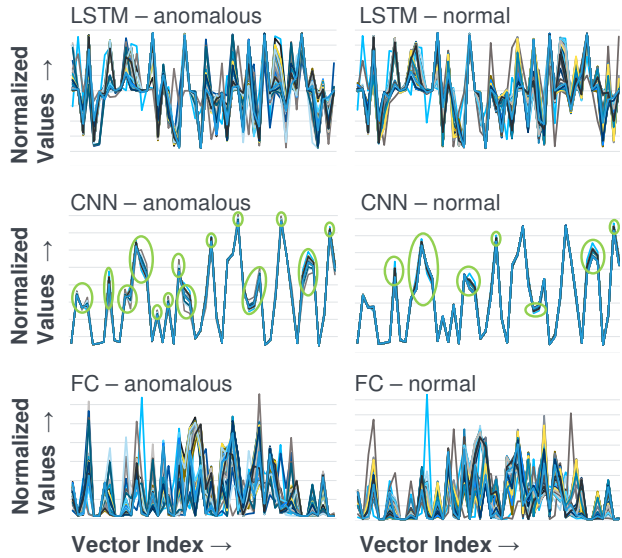

Fig. 4. 100 normalized feature vectors of code size 50 for each of the following cases: LSTM anomalous and normal (top left and right), CNN anomalous and normal (middle left and right, main deviations highlighted), FC anomalous and normal (bottom left and right)

Testing is conducted using 3,895 samples from the production of three additional products previously unseen by the algorithm. Results are depicted in Table III: The CNN-based approach outperforms the others by two orders of magnitude for a code size of 150 and order of magnitude for a code size of 50. The absolute test loss is very low (2.73x10$^{-5}$), indicating a high feature quality despite the considerable reduction of vector length (between 1:60 and 1:20).

Fig. 4 depicts exemplary feature vectors. For each of the three architectures, the same 100 normal and anomalous samples were subjected to feature extraction. The resulting feature vectors consisting of 50 normalized values each were then plotted into the diagrams. Here, the good performance of the CNN-architecture compared to that of the LSTM- or FC-architectures becomes plainly visible: Whereas the CNN-extracted feature vectors are extremely similar to each other and differ only regarding a limited number of individual features (see highlights marked with green circles), no such clear pattern emerges for the others.

### C. Modular Deep Learning Anomaly Detector

For the evaluation of both modules of the modular deep-learning-based anomaly detection algorithm, another set of experiments is conducted. Here, the pre-trained feature extraction modules of chapter IV.B are coupled with the anomaly detecting output module described in chapter III.B.

The training of the output module is conducted using different sub-sets of manually labeled data which are first processed by the different input modules. One subset of training data consists of a mixture of 10,159 normal and anomalous samples ('Mixed'). Another subset consists of 5,073 normal samples ('Normal'). The last subset consists of five copies of the 'normal' subset, resulting in 25,365 normal samples ('Normal * 5').

Table IV lists the resulting accuracies on a previously unseen test dataset of 3,384 samples, of which 1,944 are normal (57.4%), 601 anomalous (17.8%) and 839 defectives (24.8%). Again, the approach building upon the CNN-based feature extractor (FE) outperforms the other two approaches with the LSTM-based approach coming in last. Although the difference between the CNN- and FC-based approaches is only small, it must be noted that the CNN-based approach performs best on higher compressed data (vector length of 100) compared to the FC-based approach (vector length of 150). Additionally, the CNN-based approach consistently performs better on the 'normal' subset of data compared to the much larger, but not unique 'normal * 5' subset. Across all approaches, using mixed training data leads to (significantly) worse accuracies – a result in line with expectations based upon the characteristics of autoencoders.

### D. Comparison with Conventional Deep Learning Anomaly Detector

For a final evaluation of the modular deep-learning-based anomaly detection algorithm, a comparison with a conventional deep-learning-based anomaly detector is drawn. To this end, several simple FC AEs are trained and evaluated. To ensure comparability, the same hyperparameters as detailed in Table II are used. However, to ensure good performance, different numbers of layers are tried, reducing the input vector length of 3,000 to 1,500 (one layer), 1,000 (two layers) or 750 (three layers) respectively. For training, the subsets 'normal' and 'normal * 5' as described in chapter IV.C are used.

TABLE III. DIFFERENT MODULAR ANOMALY DETECTION ALGORITHMS' ACCURACIES

| FE- Architecture | | LSTM | CNN | FC |
|---|---|---|---|---|
| Vector length | Training data | | | |
| 50 | Mixed | 0.8272 | 0.8359 | 0.828 |
|  | Normal | *0.8289* | 0.9322 | 0.8301 |
|  | Normal * 5 | 0.8278 | *0.9327* | *0.835* |
| 100 | Mixed | 0.8296 | 0.8343 | 0.8292 |
|  | Normal | *0.834* | **0.9361** | 0.8666 |
|  | Normal * 5 | 0.8335 | 0.931 | *0.9108* |
| 150 | Mixed | 0.8285 | 0.8328 | 0.8286 |
|  | Normal | 0.8343 | *0.9343* | 0.8983 |
|  | Normal * 5 | **0.84** | 0.9323 | **0.9263** |

bold: architecture's best accuracy
italic: architecture's best accuracy per code size

TABLE V. CONVENTIONAL DEEP LEARNING ALGORITHM'S ACCURACIES

| Layers | Training data | Accuracy |
|---|---|---|
| 1 | Normal | 0.7805 |
| | Normal * 5 | 0.8319 |
| 2 | Normal | 0.8626 |
| | Normal * 5 | 0.9136 |
| 3 | Normal | **0.9153** |
| | Normal * 5 | 0.8554 |

bold: best accuracy

Table V lists the resulting accuracies achieved by the different versions of the FC AEs. The algorithms with two or three layers achieve results only slightly worse than those of the modular approach.

Despite the closeness of achieved accuracies, the experiment's results are clearly in favor of the modular deep-learning-based anomaly detection algorithm: Given a static feature extracting input module, the training effort for its output module is considerably less than that for the regular FC AE. This would make frequent retraining of the latter prohibitively expensive, greatly limiting its capability to adapt to changes of dynamic tasks. From our experiments, the modular approach therefore promises to be a solid foundation for transfer learning, allowing such adaptations with considerably less effort.

V. DEEP INDUSTRIAL TRANSFER LEARNING ARCHITECTURE

As outlined in chapter III, the proposed modular deep-learning-based anomaly detection algorithm shall serve as the basis for a deep transfer learning approach inspired by the dual-memory method. Fig. 5 shows such a modular deep industrial transfer learning architecture:

The pre-trained *input module* (see Fig. 5, letter A) is designed in a multi-headed manner, i.e. with a feature extraction sub-module for every sensor [28]. This allows the cooperative usage of multiple instances of the modular deep industrial transfer learning algorithm across environments with different sensor numbers and types. The resulting feature vectors can be concatenated or handled separately, depending on the scenario and the output module to be used.

The (sub-)task-specific output module (see Fig. 5, letter B) is trained using data from the respective sub-task as well as from selected previous (sub-)tasks. Due to its light-weight architecture, it is easily retrained or altogether replaced, e.g. in order to change from a regression task to a classification task.

Knowledge transfer from previous (sub-)tasks towards the current sub-task is accomplished via the *transfer module* (see Fig. 5, letter C). It stores feature vectors on all previously encountered (sub-)tasks together with corresponding meta (or context) information, e.g. regarding the sensor type, measurement time and assigned label, further describing those feature vectors and the (sub-)tasks they belong to in a representation database. In order to limit the number of feature vectors stored, only those deemed characteristic for a (sub-)task are permanently stored. The transfer module can exchange such database entries with *other learning units* (see Fig. 5, letter D).

Depending on the (sub-)task at hand, newly collected samples are then mixed with samples from previously encountered, similar (sub-)tasks to be used for (re-)training of the output module. This enlarges the (sub-)task-specific training dataset and allows a transfer of knowledge between (sub-)tasks.

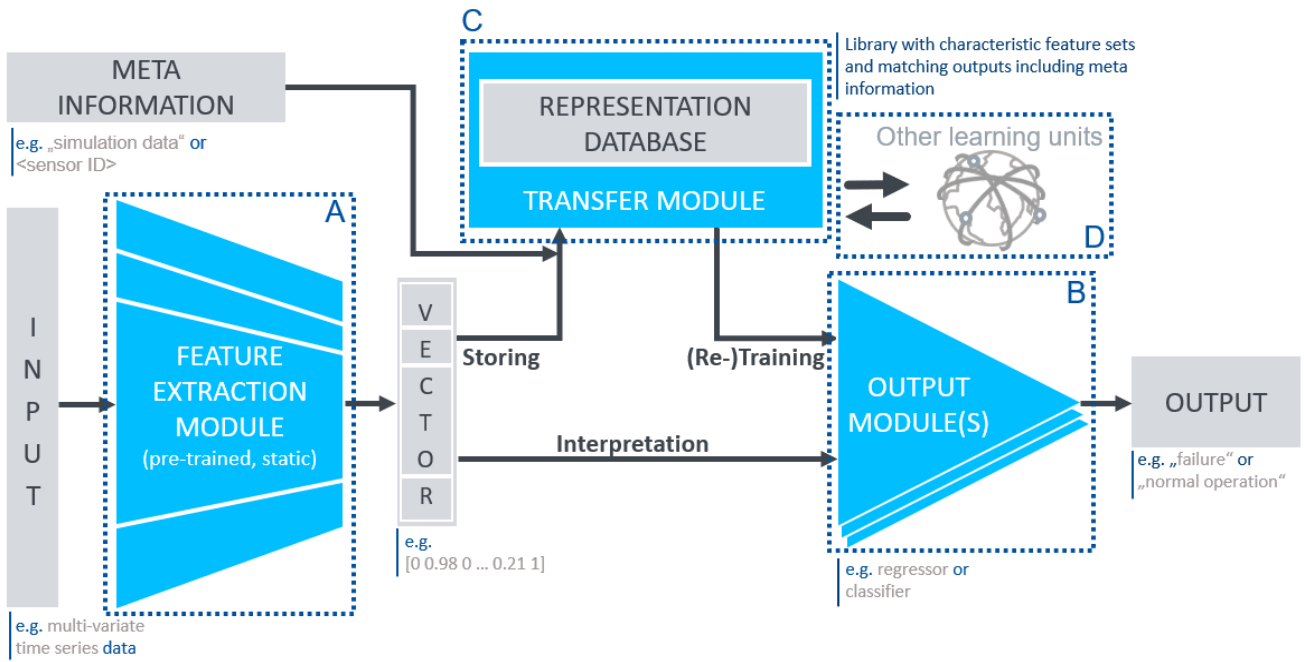

Fig. 5. Overview of a modular deep industrial transfer learning architecture

## VI. Conclusion

In this paper, a modular anomaly detection algorithm for time series data is presented and evaluated on a discrete manufacturing dataset. Special attention is given to its feasibility for future use in transfer learning:

By using separate modules for feature extraction and solving the desired task, efficient and performant machine learning across multiple related tasks as in modern image recognition shall be achieved. The feature extraction module is robustly trained on a variety of data and static thereafter. The task solving output module, here an anomaly detector, is frequently retrained to reflect changes in the dynamic task. Exchanging feature vectors of similar tasks and using them to enhance training of the output module will allow transfer learning in the future. The presented, modular anomaly detection algorithm is rigorously tested on a challenging industrial use case.

The evaluation dataset consists of multi-second events described by individual time series data. The detected anomalies are point or collective anomalies with the potential to incorporate contextual anomalies based upon the to be added transfer learning capabilities.

Our main findings are:

- Of the three examined autoencoder architectures, CNN outperform LSTM and FC neural networks in feature extraction on the dataset used.
- Using CNN, the input vector length (and thereby the amount of data to be processed by the output module) can be reduced from 3,000 to just 100 while still achieving good results.
- Using a simple two-layer FC autoencoder on the extracted feature vectors, a robust anomaly detection can be conducted – outperforming a conventional, i.e. non-modular FC-only autoencoder.

These findings underline the approach's potential for industrial transfer learning applications, i.e. applications that require frequent retrainings with only limited computational resources and training data available.

Our work will now focus on implementing and testing transfer learning capabilities based on the base approach presented in this article. Furthermore, an extensive hyperparameter optimization could further improve its performance. Concludingly, we invite other researchers to build upon this promising modular approach to mitigate the problem of dynamic tasks in industrial deep learning.

## Acknowledgment

This work was supported by the Otto Fuchs KG who provided the data.